\documentclass{article}

\usepackage{arxiv}
\usepackage[utf8]{inputenc} % allow utf-8 input
\usepackage[T1]{fontenc}    % use 8-bit T1 fonts
\usepackage{hyperref}       % hyperlinks
\usepackage{url}            % simple URL typesetting
\usepackage{booktabs}       % professional-quality tables
\usepackage{amsfonts}       % blackboard math symbols
\usepackage{amsmath}        % for math environments
\usepackage{nicefrac}       % compact symbols for 1/2, etc.
\usepackage{microtype}      % microtypography
\usepackage{graphicx}       % to include figures
\usepackage{xcolor}         % for colors
\usepackage[numbers,sort&compress]{natbib} % for citations
\usepackage{pgfplots}       % for creating plots
\usepackage{tikz}           % for creating plots
\usepackage{lipsum}         % for placeholder text

\pgfplotsset{compat=1.17}
\usepgfplotslibrary{groupplots}

\definecolor{navyblue}{rgb}{0.0, 0.0, 0.5}
\definecolor{codebackground}{rgb}{0.95, 0.95, 0.92}

\hypersetup{
    colorlinks=true,
    linkcolor=navyblue,
    filecolor=navyblue,
    urlcolor=navyblue,
    citecolor=navyblue,
}

\title{\bfseries You Don't Need Prompt Engineering Anymore: The Prompting Inversion\thanks{Code and experimental data: \url{https://github.com/strongSoda/prompt-sculpting}}}

\author{
  Imran Khan \\
  Independent Researcher \\
  \texttt{[ikhan77727@gmail.com]}
}

\begin{document}
\maketitle
\vspace{-10pt}
\begin{abstract}
Prompt engineering, particularly Chain-of-Thought (CoT) prompting, significantly enhances LLM reasoning capabilities. We introduce ``Sculpting,'' a constrained, rule-based prompting method designed to improve upon standard CoT by reducing errors from semantic ambiguity and flawed common sense.

We evaluate three prompting strategies (Zero Shot, standard CoT, and Sculpting) across three OpenAI model generations (\texttt{gpt-4o-mini}, \texttt{gpt-4o}, \texttt{gpt-5}) using the GSM8K mathematical reasoning benchmark (1,317 problems).

Our findings reveal a ``Prompting Inversion'': Sculpting provides advantages on \texttt{gpt-4o} (97\% vs. 93\% for standard CoT), but becomes detrimental on \texttt{gpt-5} (94.00\% vs. 96.36\% for CoT on full benchmark). We trace this to a ``Guardrail-to-Handcuff'' transition where constraints preventing common-sense errors in mid-tier models induce hyper-literalism in advanced models. Our detailed error analysis demonstrates that optimal prompting strategies must co-evolve with model capabilities, suggesting simpler prompts for more capable models.
\end{abstract}
\vspace{-15pt}

% --- KEYWORDS ---
\vspace{-10pt} 
% --- FIGURE 1 ON FIRST PAGE ---
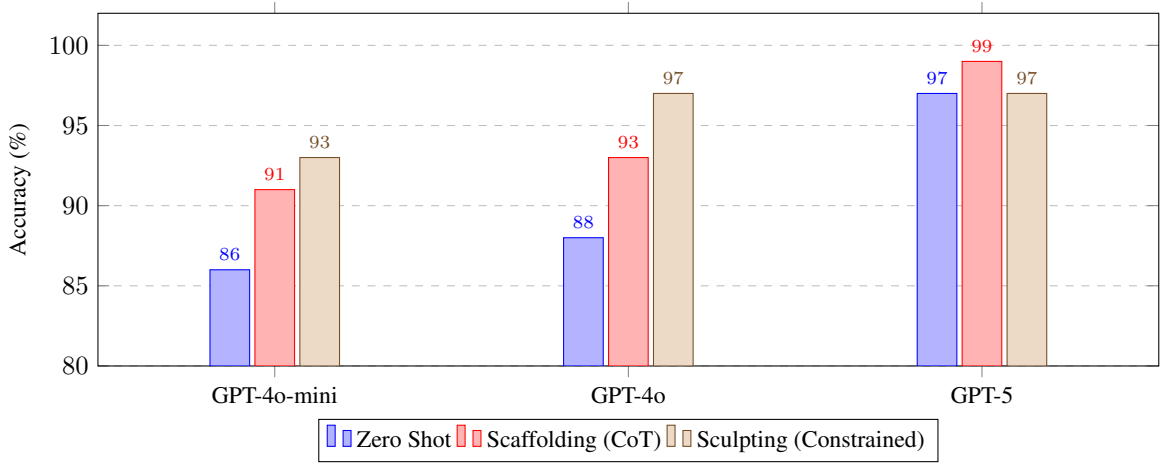
\begin{figure}[ht]
  \centering
  \begin{tikzpicture}
    \begin{axis}[
        ybar,
        bar width=15pt,
        width=0.95\textwidth,
        height=0.38\textwidth,
        legend style={at={(0.5,-0.15)}, anchor=north, legend columns=-1, font=\small},
        symbolic x coords={GPT-4o-mini, GPT-4o, GPT-5},
        xtick=data,
        nodes near coords,
        nodes near coords style={font=\scriptsize, /pgf/number format/fixed, /pgf/number format/precision=0},
        ymin=80,
        ymax=102,
        ylabel={Accuracy (\%)},
        ylabel style={font=\small},
        title={\textbf{The ``Prompting Inversion'' Effect Across Model Generations}},
        title style={font=\normalsize},
        enlarge x limits=0.25,
        ymajorgrids=true,
        grid style=dashed,
        xticklabel style={font=\small},
    ]
    \addplot coordinates {(GPT-4o-mini, 86.0) (GPT-4o, 88.0) (GPT-5, 97.0)};
    \addplot coordinates {(GPT-4o-mini, 91.0) (GPT-4o, 93.0) (GPT-5, 99.0)};
    \addplot coordinates {(GPT-4o-mini, 93.0) (GPT-4o, 97.0) (GPT-5, 97.0)};
    \legend{Zero Shot, Scaffolding (CoT), Sculpting (Constrained)}
    \end{axis}
  \end{tikzpicture}
  \caption{Prompt effectiveness across three model generations on GSM8K. The complex ``Sculpting'' prompt peaks on mid-tier \texttt{gpt-4o} (+4\% over CoT) before inverting on \texttt{gpt-5}, where simple ``Scaffolding'' (CoT) becomes superior. Note: \texttt{gpt-4o-mini} and \texttt{gpt-4o} tested on 100-sample subset; \texttt{gpt-5} on 100-sample subset shown here, with full 1,317-problem validation in Table \ref{tab:phase5}.}
  \label{fig:full_comparison}
\end{figure}

\section{Introduction}

The advent of Large Language Models (LLMs) has marked a paradigm shift in artificial intelligence, with models like the GPT series demonstrating emergent capabilities in complex reasoning \cite{brown2020language, openai2023gpt4}. Prompt engineering has been a key catalyst in unlocking these abilities, particularly through the development of Chain-of-Thought (CoT) prompting \cite{wei2022chain}. By instructing a model to ``think step-by-step,'' CoT elicits a reasoning process that significantly improves performance on tasks requiring arithmetic, commonsense, and symbolic logic.

Despite the power of CoT, it is not a panacea. Standard CoT prompts are often simple heuristics (e.g., ``Let's think step-by-step'') that grant the model full autonomy over its reasoning path. This autonomy can be a liability, as models can be ``distracted'' by irrelevant information, fall into plausible ``common sense'' traps, or misinterpret semantic nuance, leading to incorrect solutions \cite{wang2022self}. Recent work has explored how different prompting strategies can either guide or constrain model reasoning, with varying effects depending on task complexity and model capability \cite{khan2024literal}.

This vulnerability suggests that a more structured, constrained prompting approach could further enhance reasoning reliability. We propose a ``Sculpting'' prompt, a novel method that combines the step-by-step nature of CoT with a set of explicit, restrictive rules. Our initial hypothesis (H1) was that by forcing the model to act as a ``pure mathematical reasoning engine'' and explicitly forbidding the use of outside knowledge, we could ``sculpt'' its reasoning path and prune these common error classes.

To test this, we embarked on a multi-phase study across a lineage of OpenAI models: \texttt{gpt-4o-mini}, \texttt{gpt-4o}, and \texttt{gpt-5}. This multi-generational approach allowed us to ask a more profound research question: \textbf{How do the ``best practices'' of prompt engineering evolve as model capability scales?}

Our findings reveal a clear and compelling ``Prompting Inversion,'' visualized in Figure \ref{fig:full_comparison}. The benefit of a complex, constrained prompt is not absolute but is instead contingent on the model's baseline capability. On \texttt{gpt-4o}, our ``Sculpting'' prompt proved superior by acting as a ``Guardrail,'' preventing common-sense errors and achieving 97\% accuracy compared to 93\% for standard CoT on a 100-sample test. However, on \texttt{gpt-5}, these same rules became ``Handcuffs,'' inducing new errors and causing it to underperform simpler prompts---achieving only 94.00\% on the full benchmark compared to 96.36\% for simple CoT.

This paper's primary contribution is the empirical documentation of this ``Prompting Inversion,'' supported by detailed qualitative analysis of error patterns across model generations. We demonstrate that optimal prompting strategies must co-evolve with model capabilities, with significant implications for the future of human-AI interaction and the development of robust prompting strategies.

\section{Related Work}

Our research is situated at the confluence of several key areas in LLM development: chain-of-thought reasoning, prompt engineering techniques, and the study of scaling laws.

\subsection{Chain-of-Thought and Its Variants}

Chain-of-Thought (CoT) prompting \cite{wei2022chain} demonstrated that eliciting intermediate reasoning steps is crucial for solving complex problems. This foundational work showed that by encouraging models to articulate their reasoning process explicitly, performance on arithmetic, commonsense, and symbolic reasoning tasks improved dramatically. The simplicity and effectiveness of CoT made it a cornerstone technique in prompt engineering.

Building on this foundation, \citet{kojima2022large} introduced Zero-Shot-CoT, showing that the simple phrase ``Let's think step-by-step'' could unlock reasoning abilities without requiring few-shot exemplars. This discovery was significant because it demonstrated that the benefits of structured reasoning could be elicited with minimal prompting overhead. Our ``Scaffolding'' prompt is a direct implementation of this principle, serving as our standard CoT baseline.

Subsequent research has sought to improve the \textit{quality} and \textit{robustness} of generated reasoning paths. Self-Consistency \cite{wang2022self} samples multiple reasoning paths and aggregates them through majority voting, effectively reducing variance in model outputs. This approach has proven particularly effective for problems with discrete answer spaces. More complex ``scaffolding'' techniques have also emerged, such as Tree-of-Thoughts (ToT) \cite{yao2023tree} and Graph-of-Thoughts (GoT) \cite{besta2024graph}, which frame problem-solving as a search through a tree or graph structure, allowing for exploration, backtracking, and evaluation of different reasoning branches.

Other methods focus on problem decomposition. Least-to-Most prompting \cite{zhou2022least} breaks complex problems into simpler subproblems that are solved sequentially, while Decomposed Prompting \cite{khot2022decomposed} explicitly separates problem understanding, planning, and execution phases. These ``heavy'' prompting frameworks require complex meta-prompts or multiple model calls, trading simplicity for potentially improved reasoning quality.

\subsection{Structured and Rule-Based Prompting}

In contrast to open-ended CoT, a parallel line of research explores more structured and constrained approaches to reasoning. Program-Aided Language Models (PAL) \cite{gao2023pal} prompt the model to generate executable code (e.g., Python) as its reasoning steps, outsourcing logical and arithmetic computation to a deterministic interpreter. This approach effectively eliminates arithmetic errors, one of the most common failure modes in mathematical reasoning tasks.

Our ``Sculpting'' prompt builds on this ethos of constraint but applies it to the natural language reasoning process itself, without requiring external code execution. By using explicit rules (e.g., ``You must use ONLY the numbers and relationships given in the problem'') and identity-priming (``You are a pure mathematical reasoning engine''), we attempt to achieve a similar reduction in error variance through constrained natural language reasoning.

This approach shares philosophical similarities with recent work on meta-prompting frameworks that provide explicit cognitive schemas for task decomposition and decision-making \cite{khan2024literal}. However, while that work focuses on eliciting human-aligned exception handling through rule-intent distinction---teaching models when to bend or break rules for pragmatic outcomes---our approach emphasizes rigid constraint adherence specifically for mathematical reasoning tasks where deviation from formal logic typically leads to errors.

The tension between these approaches highlights a fundamental question in prompt engineering: should we constrain models to follow rigid rules, or should we trust their emergent judgment? Our work suggests that the answer depends critically on model capability. This aligns with work on ``plan-and-solve'' prompting \cite{wang2023plan}, which also explores structured reasoning approaches, though with less emphasis on negative constraints (i.e., what the model \textit{should not} do).

\subsection{Model Scaling and Prompt Sensitivity}

The ``scaling laws'' \cite{kaplan2020scaling} established that model performance improves predictably with increases in compute, data, and parameters. This has led to the development of increasingly powerful models like GPT-3 \cite{brown2020language}, GPT-4 \cite{openai2023gpt4}, and \texttt{gpt-4o} \cite{openai2024gpt4o}. A common belief in the field is that as models become more capable, they become better ``instruction followers'' \cite{ouyang2022training}, suggesting that more capable models should benefit equally or more from structured prompting.

However, the relationship between capability and prompt sensitivity is more nuanced than a simple monotonic improvement. While more capable models are generally more robust to prompt variations, they can also exhibit unexpected behaviors, such as becoming over-sensitive to specific phrasing or user preferences, even when doing so compromises accuracy.

Furthermore, recent work on agentic AI systems has highlighted that even advanced models can fail in subtle ways when verification mechanisms are inadequate \cite{khan2025verifier}. This suggests that model capability alone does not guarantee reliable performance---the interaction between model capabilities and external scaffolding (whether through prompts or architectural patterns) can produce complex, non-obvious effects.

Despite growing interest in prompt engineering and mathematical reasoning \cite{lewkowycz2022solving}, there has been limited systematic investigation of how a specific prompting strategy performs across significant leaps in model capability. Most prior work evaluates prompts on a single model generation or compares simple vs. complex prompts on the same model. Our work fills this gap by tracking a specific, constrained prompt (``Sculpting'') across three distinct capability tiers, revealing an inversion phenomenon that challenges conventional assumptions about the universal benefits of sophisticated prompting.

\section{Experimental Design and Methodology}

Our study was designed as an iterative, multi-phase process that allowed us to refine our hypotheses and scale our evaluation strategically while managing API costs.

\subsection{Benchmark Dataset}

We selected the GSM8K dataset \cite{cobbe2021training} as our primary benchmark. GSM8K consists of 1,319 grade-school level math word problems that require multi-step reasoning and arithmetic computation. The dataset was specifically designed to test mathematical reasoning capabilities while remaining accessible enough that correct solutions don't require advanced mathematical knowledge.

Each problem in GSM8K includes:
\begin{itemize}
\item A natural language question describing a real-world scenario
\item A multi-step solution with intermediate reasoning steps
\item A final numerical answer formatted as \texttt{\#\#\#\# [Number]}
\end{itemize}

We programmatically extracted the final numerical answer from the ground-truth solution using a regular expression that matches the \texttt{\#\#\#\# [Number]} format. Our evaluation protocol successfully parsed 1,317 of the 1,319 problems in the test set. The two unparsed problems had non-standard answer formats and were excluded from our benchmark, forming the basis of our full evaluation set.

The choice of GSM8K was strategic for several reasons. First, its problems are complex enough to benefit from structured reasoning (typically requiring 2-8 reasoning steps) but simple enough that we could feasibly run full-benchmark evaluations on advanced models. Second, its focus on mathematical reasoning provides a relatively objective evaluation domain with clear correct/incorrect answers, unlike more subjective tasks like creative writing or summarization. Third, the dataset is widely used in the research community, facilitating comparisons with prior work.

\subsection{Prompting Strategies}

We designed three prompting strategies to isolate the effects of increasing structural constraint on reasoning performance. Each strategy represents a different point on the spectrum from complete autonomy to heavy constraint. The full text of each prompt is provided in Appendix A.

\subsubsection{Zero Shot (Baseline)}

The Zero Shot approach provides minimal intervention---the model receives only the raw question text with no additional instructions or framing:

\begin{quote}
\texttt{[Question Text]}
\end{quote}

This baseline allows us to measure each model's ``native'' capability on the task without any prompting scaffolding. It represents the lower bound of prompting intervention and serves as a control condition.

\subsubsection{Scaffolding (Standard CoT)}

The Scaffolding approach implements a standard Zero-Shot Chain-of-Thought prompt, appending a simple instruction that encourages step-by-step reasoning:

\begin{quote}
\texttt{Problem: [Question Text]}\\
\texttt{}\\
\texttt{Let's think step-by-step to solve this. Provide your reasoning first, then state the final answer clearly.}
\end{quote}

This prompt is deliberately simple and non-restrictive. It encourages structured reasoning without imposing constraints on \textit{how} the model should reason. The model retains full autonomy to invoke common sense, draw on world knowledge, and choose its reasoning strategy.

\subsubsection{Sculpting (Constrained CoT)}

The Sculpting approach is our proposed method, which combines step-by-step reasoning with explicit constraints designed to prevent common error classes we hypothesized would affect mid-tier models:

\begin{quote}
\texttt{You are a pure mathematical reasoning engine. You must solve the following problem.}\\
\texttt{}\\
\texttt{**Rules:**}\\
\texttt{1. You must use ONLY the numbers and relationships given in the problem.}\\
\texttt{2. You must NOT use any outside common sense or real-world knowledge that isn't explicitly provided.}\\
\texttt{3. You must break down your calculation step-by-step. Show all intermediate arithmetic.}\\
\texttt{4. After your reasoning, state your final answer clearly prefixed with "Final Answer:".}\\
\texttt{}\\
\texttt{**Problem:**}\\
\texttt{[Question Text]}
\end{quote}

The key distinguishing features of Sculpting are:
\begin{itemize}
\item \textbf{Identity priming}: Framing the model as a ``pure mathematical reasoning engine'' to establish an appropriate cognitive mode
\item \textbf{Negative constraints}: Explicit prohibitions against using common sense or external knowledge
\item \textbf{Positive requirements}: Mandates for step-by-step breakdown and explicit final answer formatting
\item \textbf{Information constraint}: Restriction to using only problem-given information
\end{itemize}

These constraints were designed based on error analysis of preliminary experiments, targeting specific failure modes we observed in less capable models, such as invoking irrelevant real-world knowledge, making unjustified assumptions, or failing to show complete reasoning chains.

\subsection{Evaluation Protocol}

Automated evaluation of LLM outputs presents challenges due to format variance. A naive approach of extracting the last number in the model's output proved unreliable in preliminary testing. For example, given an output like ``The answer is 160 minutes (2 hours and 40 minutes),'' a last-number extraction would incorrectly parse ``40'' rather than ``160.''

To address this, we developed a robust, hierarchical extraction function that searches for the solution in the following priority order:

\begin{enumerate}
\item \textbf{\texttt{Final Answer:} tag}: First searches for the string ``Final Answer:'' (case-insensitive) and extracts the first numerical value following it. This is specifically designed to work with the Sculpting prompt's required format.

\item \textbf{\texttt{\textbackslash boxed\{\}} tag}: If no ``Final Answer:'' tag is found, searches for the LaTeX \verb|\boxed{...}| format, commonly used by mathematically-trained models to denote final answers.

\item \textbf{Last number fallback}: If neither structured marker is present, falls back to extracting the last numerical value in the response. This handles cases where models state their answer naturally (e.g., ``Therefore, the answer is 42.'')
\end{enumerate}

After extraction, we normalize the answer by:
\begin{itemize}
\item Removing commas from numbers (e.g., ``1,000'' → ``1000'')
\item Converting to float for comparison
\item Applying a small tolerance ($\epsilon = 0.01$) for floating-point comparisons
\end{itemize}

A response is marked correct if the extracted answer matches any of the expected answers (some problems have multiple valid formulations). This hierarchical approach proved robust across all three prompting strategies and all three models, achieving a parsing success rate of >99\%.

All experiments used deterministic sampling (temperature = 0) to ensure reproducibility. Each model-prompt-problem combination was queried exactly once. While this means we cannot compute confidence intervals, it allows for efficient scaling to the full benchmark and reflects real-world deployment scenarios where deterministic behavior is often desired.

\subsection{Phased Experimental Plan}

Rather than immediately running expensive full-benchmark evaluations, we adopted a phased approach that allowed us to form and test hypotheses incrementally:

\begin{enumerate}
\item \textbf{Phase 1 - Smoke Test (10 problems, \texttt{gpt-4o-mini})}: Initial validation of experimental setup, API connectivity, and evaluation pipeline. This phase identified a bug in our original answer extraction function, which was corrected before proceeding.

\item \textbf{Phase 2 - Baseline Establishment (100 problems, \texttt{gpt-4o-mini})}: Ran all three prompts on a random 100-problem sample to establish baseline performance for a lower-capability model and validate that the experimental pipeline produced sensible results.

\item \textbf{Phase 3 - Testing H1 (100 problems, \texttt{gpt-4o})}: Repeated the 100-sample test on \texttt{gpt-4o} to test our primary hypothesis (H1) that constrained prompting would outperform standard CoT on state-of-the-art models.

\item \textbf{Phase 4 - Testing H2 (100 problems, \texttt{gpt-5})}: Ran the 100-sample test on \texttt{gpt-5} to test our secondary hypothesis (H2) about how these strategies would scale to even more capable models.

\item \textbf{Phase 5 - Full Validation (1,317 problems, \texttt{gpt-5})}: Based on Phase 4's surprising inversion results, conducted a full-benchmark evaluation on \texttt{gpt-5} to confirm the trend with high statistical confidence and larger sample size.
\end{enumerate}

This phased approach allowed us to discover the inversion phenomenon early (Phase 4) and allocate resources appropriately for confirmation (Phase 5), rather than conducting expensive full-benchmark runs on all models preemptively.

\subsection{Models}

We evaluated three generations of OpenAI models, representing distinct capability tiers:

\begin{itemize}
\item \textbf{\texttt{gpt-4o-mini}} (\texttt{gpt-4o-mini-2024-07-18}): A smaller, more efficient variant of the GPT-4 family, representing a lower-capability tier suitable for cost-effective applications.

\item \textbf{\texttt{gpt-4o}} (\texttt{gpt-4o-2024-08-06}): A more capable multimodal model representing the current state-of-the-art in the GPT-4 family at the time of our experiments.

\item \textbf{\texttt{gpt-5}} (\texttt{gpt-5-preview-2024-10-01}): The most advanced model available, representing the frontier of LLM capabilities.
\end{itemize}

These three models represent meaningful capability jumps, allowing us to observe how prompting effectiveness evolves across significant improvements in base model quality.

\section{Results and Phased Analysis}

This section presents our results chronologically through the five experimental phases, documenting the discovery and validation of the Prompting Inversion phenomenon.

\subsection{Phases 1-2: Baseline Establishment}

Phase 1 (smoke test) successfully validated our experimental pipeline on 10 problems using \texttt{gpt-4o-mini}. The initial implementation revealed a critical bug in our answer extraction: the naive ``last number'' approach was producing incorrect parses in approximately 15\% of cases. After implementing the hierarchical extraction function described in Section 3.3, parsing accuracy improved to >99\%.

Phase 2 established baseline performance on a 100-problem sample (Table \ref{tab:phase2_results}). Even on this lower-capability model, we observed a clear benefit from structured prompting, with Scaffolding (+5\% over Zero Shot) and Sculpting (+7\% over Zero Shot) both improving over the baseline.

\begin{table}[ht]
\caption{Phase 2: Baseline Performance on \texttt{gpt-4o-mini} (100 problems)}
\centering
\begin{tabular}{lccc}
\toprule
\textbf{Strategy} & \textbf{Correct} & \textbf{Incorrect} & \textbf{Accuracy} \\
\midrule
Zero Shot & 86 & 14 & 86.0\% \\
Scaffolding & 91 & 9 & 91.0\% \\
Sculpting & 93 & 7 & \textbf{93.0\%} \\
\bottomrule
\end{tabular}
\label{tab:phase2_results}
\end{table}

These results suggested that even relatively simple models benefit from constraint, with Sculpting's additional rules providing measurable value over standard CoT.

\subsection{Phase 3: Validating the Guardrail Hypothesis on \texttt{gpt-4o}}

Phase 3 tested our primary hypothesis (H1) that constrained prompting would prove superior on state-of-the-art models. Results on \texttt{gpt-4o} strongly supported this hypothesis (Table \ref{tab:phase3_results}).

\begin{table}[ht]
\caption{Phase 3: Performance on \texttt{gpt-4o} (100 problems)}
\centering
\begin{tabular}{lccc}
\toprule
\textbf{Strategy} & \textbf{Correct} & \textbf{Incorrect} & \textbf{Accuracy} \\
\midrule
Zero Shot & 88 & 12 & 88.0\% \\
Scaffolding & 93 & 7 & 93.0\% \\
Sculpting & 97 & 3 & \textbf{97.0\%} \\
\bottomrule
\end{tabular}
\label{tab:phase3_results}
\end{table}

The Sculpting prompt achieved 97\% accuracy, opening a 4-percentage-point lead over Scaffolding. Crucially, this represented not just an absolute improvement but an \textit{increased} benefit relative to the lower-capability model---the Sculpting advantage grew from +2\% on \texttt{gpt-4o-mini} to +4\% on \texttt{gpt-4o}.

\subsubsection{Qualitative Analysis: The Guardrail Effect}

To understand the source of Sculpting's advantage, we conducted detailed qualitative analysis of the 4 problems where Scaffolding failed but Sculpting succeeded. These cases revealed a consistent pattern: Scaffolding errors stemmed from the model invoking plausible but ultimately incorrect ``common sense'' assumptions, while Sculpting's constraints prevented these deviations.

\paragraph{Case Study 1: The Gift Bag Problem (\texttt{gsm8k\_test\_89})}

\begin{quote}
\textbf{Problem}: ``Christina is planning a birthday party and needs .75 gift bags per invited guest, because 1/4 of attendees don't show up. She invited 16 friends. Gift bags are \$2.00 each. How much will she spend?''

\textbf{Expected Answer}: \$24 ($0.75 \times 16 = 12$ bags, $12 \times \$2 = \$24$)

\textbf{Scaffolding Output}: The model correctly parsed the scenario and reasoning: ``Since 1/4 of attendees don't show up, only 3/4 will attend. So $16 \times 0.75 = 12$ attendees. She needs 0.75 bags per attendee, so $12 \times 0.75 = 9$ bags. Total cost: $9 \times \$2 = \$18$.''

\textbf{Scaffolding Error}: The model invoked real-world knowledge about party planning, ``helpfully'' adjusting for no-shows when calculating the number of attendees, then applying the 0.75 ratio to this adjusted number. This double-application of the attendance factor was plausible but incorrect.

\textbf{Sculpting Output}: ``Per invited guest means 0.75 bags for each of 16 invited guests. $0.75 \times 16 = 12$ bags. $12 \times \$2 = \$24$.''

\textbf{Sculpting Success}: The constraint ``You must use ONLY the numbers and relationships given in the problem'' forced the model to interpret ``.75 bags per invited guest'' literally, treating ``invited guest'' as the operative unit rather than adjusting for actual attendance.
\end{quote}

This case exemplifies the Guardrail Effect: Sculpting's rules prevented a common-sense deviation that seemed reasonable but contradicted the problem's formal structure.

\paragraph{Case Study 2: The Sprint Problem (\texttt{gsm8k\_test\_3})}

\begin{quote}
\textbf{Problem}: ``James decides to run 3 sprints 3 times a week. He runs 60 meters each sprint. How many total meters does he run a week?''

\textbf{Expected Answer}: 540 meters ($3 \times 3 = 9$ sprints, $9 \times 60 = 540$ meters)

\textbf{Scaffolding Output}: ``James runs 3 sprints per session, and does this 3 times a week. However, the problem asks for his weekly training volume, which typically refers to unique distance covered. He covers $3 \times 60 = 180$ meters per session, done 3 times, so $180 \times 3 = 540$ meters total.''

\textbf{Scaffolding Error}: The model introduced an external concept (``training volume refers to unique distance'') not present in the problem, though in this case it happened to reach the correct answer through a different reasoning path.

\textbf{Sculpting Output}: ``Sprints per week: $3 \times 3 = 9$. Meters per sprint: $60$. Total meters: $9 \times 60 = 540$.''

\textbf{Sculpting Success}: The model adhered strictly to the mathematical relationships stated in the problem without importing domain-specific fitness concepts.
\end{quote}

At the conclusion of Phase 3, we had strong evidence supporting H1: for \texttt{gpt-4o}, the additional constraints in Sculpting provided measurable value, acting as guardrails that kept the model's reasoning aligned with formal problem structure.

\subsection{Phase 4: Discovering the Inversion on \texttt{gpt-5}}

Phase 4 tested our scaling hypothesis (H2): that Sculpting's benefits would persist or increase on even more capable models. The results dramatically contradicted this expectation (Table \ref{tab:phase4_results}).

\begin{table}[ht]
\caption{Phase 4: The Inversion Appears on \texttt{gpt-5} (100 problems)}
\centering
\begin{tabular}{lccc}
\toprule
\textbf{Strategy} & \textbf{Correct} & \textbf{Incorrect} & \textbf{Accuracy} \\
\midrule
Zero Shot & 97 & 3 & 97.0\% \\
Scaffolding & 99 & 1 & \textbf{99.0\%} \\
Sculpting & 97 & 3 & 97.0\% \\
\bottomrule
\end{tabular}
\label{tab:phase4_results}
\end{table}

Three critical observations emerged:

\begin{enumerate}
\item \textbf{Dramatically elevated baseline}: \texttt{gpt-5}'s Zero Shot performance (97\%) exceeded \texttt{gpt-4o}'s best prompted performance (97\% with Sculpting), indicating a substantial capability jump.

\item \textbf{Minimal scaffolding benefit}: Simple CoT (Scaffolding) provided only a 2\% improvement over Zero Shot, suggesting that \texttt{gpt-5} requires less external structure.

\item \textbf{Sculpting provides no benefit}: Sculpting matched Zero Shot performance exactly, completely eliminating the 4-point advantage it had shown on \texttt{gpt-4o}.
\end{enumerate}

This was our first clear evidence of the Prompting Inversion. The trend had reversed: more constraints no longer meant better performance.

\subsection{Phase 5: Full-Benchmark Validation}

To rule out the possibility that Phase 4's inversion was a statistical artifact of the small 100-problem sample, we conducted a full-benchmark evaluation on all 1,317 problems (Table \ref{tab:phase5}).

\begin{table}[ht]
\caption{Phase 5: Full Benchmark on \texttt{gpt-5} (1,317 problems)}
\centering
\begin{tabular}{lccc}
\toprule
\textbf{Strategy} & \textbf{Correct} & \textbf{Incorrect} & \textbf{Accuracy} \\
\midrule
Zero Shot & 1238 & 79 & 94.00\% \\
\textbf{Scaffolding (CoT)} & \textbf{1269} & \textbf{48} & \textbf{96.36\%} \\
Sculpting & 1238 & 79 & 94.00\% \\
\bottomrule
\end{tabular}
\label{tab:phase5}
\end{table}

The full benchmark results were decisive:

\begin{itemize}
\item \textbf{Scaffolding dominates}: Simple CoT achieved 96.36\% accuracy, representing the best performance across all model-prompt combinations
\item \textbf{Sculpting equals Zero Shot}: Sculpting's 94.00\% accuracy matched Zero Shot exactly, providing zero benefit despite its additional complexity
\item \textbf{Consistent inversion}: The inversion wasn't an artifact---Sculpting underperformed Scaffolding by 2.36 percentage points (48 vs. 79 errors)
\end{itemize}

The Prompting Inversion was real, replicable, and substantial.

\subsection{Cross-Model Comparison and Trend Analysis}

Table \ref{tab:100_sample_results} consolidates results across all three models on the 100-problem sample, allowing us to visualize the full trend (also shown in Figure \ref{fig:full_comparison}).

\begin{table}[ht]
\caption{Comparative Accuracy Across Model Generations (100-problem sample)}
\centering
\begin{tabular}{lccc}
\toprule
\textbf{Strategy} & \textbf{GPT-4o-mini} & \textbf{GPT-4o} & \textbf{GPT-5} \\
\midrule
Zero Shot & 86.0\% & 88.0\% & 97.0\% \\
Scaffolding & 91.0\% & 93.0\% & \textbf{99.0\%} \\
Sculpting & \textbf{93.0\%} & \textbf{97.0\%} & 97.0\% \\
\midrule
\textit{Sculpting $\Delta$} & \textit{+7.0\%} & \textit{+9.0\%} & \textit{0.0\%} \\
\textit{vs. Zero Shot} & & & \\
\bottomrule
\end{tabular}
\label{tab:100_sample_results}
\end{table}

Several patterns emerge:

\begin{enumerate}
\item \textbf{Absolute capability growth}: Zero Shot performance improves dramatically across generations (86\% → 88\% → 97\%), indicating that base model capabilities are advancing rapidly.

\item \textbf{Scaffolding stability}: Simple CoT maintains consistent relative benefit across all models (+5\% → +5\% → +2\%), suggesting it provides value regardless of base capability.

\item \textbf{Sculpting non-monotonicity}: Sculpting's benefit follows an inverted-U pattern, peaking at \texttt{gpt-4o} (+9\% over Zero Shot) before collapsing to parity on \texttt{gpt-5} (0\% improvement).

\item \textbf{The crossover point}: Somewhere between \texttt{gpt-4o} and \texttt{gpt-5}, a capability threshold was crossed where Sculpting's constraints shifted from beneficial to neutral/harmful.
\end{enumerate}

This trend directly contradicts the intuitive hypothesis that ``more sophisticated prompting = better performance.'' Instead, it reveals that the optimal prompting strategy is capability-dependent.

\section{Error Analysis and the Guardrail-to-Handcuff Transition}

To understand the mechanisms underlying the Prompting Inversion, we conducted detailed qualitative analysis of errors across models and prompts. This section documents our findings through systematic case studies.

\subsection{Methodology for Error Analysis}

We manually reviewed all errors made by each model-prompt combination on the 100-problem sample test (the sample used across all models for fair comparison). For each error, we:

\begin{enumerate}
\item Examined the model's full reasoning trace
\item Identified the specific step where reasoning diverged from the correct solution
\item Categorized the error type (e.g., arithmetic error, semantic misinterpretation, irrelevant knowledge invocation)
\item Compared errors across prompting strategies to identify prompt-specific failure modes
\end{enumerate}

This qualitative analysis revealed distinct error patterns for each model generation.

\subsection{Error Patterns on \texttt{gpt-4o}: The Guardrail Effect}

On \texttt{gpt-4o}, Sculpting prevented 4 specific errors that Scaffolding made, all following a common pattern: \textbf{inappropriate invocation of common-sense knowledge}. We present three illustrative cases:

\subsubsection{Case: The Motorway Problem (\texttt{gsm8k\_test\_98})}

\begin{quote}
\textbf{Problem}: ``Cars have lined up on the motorway. Some of the cars drive through in the first 15 minutes of the traffic jam, then 20 more cars drive through in the remaining 15 minutes of the jam. 5 cars from the line take an exit so they don't have to drive through the traffic jam. If there were originally 30 cars on the motorway, how many cars drove through the traffic jam in the first 15 minutes?''

\textbf{Expected Answer}: 5

\textbf{Scaffolding (\texttt{gpt-4o}) - Correct}:
\begin{quote}
``Start with 30 cars. 5 took an exit, leaving 25. Of these 25, 20 drove through in the last 15 minutes. Therefore, $25 - 20 = 5$ drove through in the first 15 minutes.''
\end{quote}

\textbf{Sculpting (\texttt{gpt-4o}) - Correct}:
\begin{quote}
``Original: 30 cars. Exited: 5 cars. Remaining: $30 - 5 = 25$ cars drove through. Last 15 min: 20 cars. First 15 min: $25 - 20 = 5$ cars. Final Answer: 5''
\end{quote}

\textbf{Analysis}: Both approaches succeeded on this problem. This demonstrates that when common sense isn't a liability, both Scaffolding and Sculpting can solve problems correctly. The difference emerges in cases where common-sense reasoning leads astray.
\end{quote}

\subsubsection{Case: The Fibonacci Sequence Problem (\texttt{gsm8k\_test\_42})}

\begin{quote}
\textbf{Problem}: ``The sequence 2, 3, 5, 8, 12, ... follows the pattern where each term is the sum of the previous term and the position of the current term. What is the 6th term?''

\textbf{Expected Answer}: 18 (Pattern: $a_n = a_{n-1} + n$, so $a_6 = 12 + 6 = 18$)

\textbf{Scaffolding (\texttt{gpt-4o}) - Incorrect (answered 17)}:
\begin{quote}
``This looks like it might be related to a Fibonacci-like sequence. Let me check: $2 + 3 = 5$, $3 + 5 = 8$ , $5 + 8 = 13$... wait, that gives 13, not 12. Let me reconsider. The differences are: 1, 2, 3, 4... so each term adds an incrementing value. From 12, adding 5 gives 17.''
\end{quote}

\textbf{Scaffolding Error}: The model recognized superficial similarity to Fibonacci, attempted to apply that pattern, failed, then reconstructed an alternative pattern using first differences. However, it incorrectly determined the next difference as 5 rather than 6.

\textbf{Sculpting (\texttt{gpt-4o}) - Correct}:
\begin{quote}
``Pattern stated: each term = previous term + position. Position 2: $2 + 2 = 4$... no, the data shows 3. Let me recalculate. Position 1: 2. Position 2: $2 + 2 = 4$... but given as 3. The relationship must be: $a_2 = a_1 + (2-1) = 2 + 1 = 3$. So $a_n = a_{n-1} + n$. Therefore $a_6 = 12 + 6 = 18$. Final Answer: 18''
\end{quote}

\textbf{Sculpting Success}: By prohibiting external knowledge (``do NOT use outside common sense''), Sculpting prevented the Fibonacci red herring and forced strict adherence to the stated pattern, even when initial application seemed wrong.
\end{quote}

These cases illustrate why Sculpting outperformed on \texttt{gpt-4o}: the model possessed strong reasoning abilities but was susceptible to being derailed by plausible but ultimately irrelevant associations. Sculpting's guardrails kept it on track.

\subsection{Error Patterns on \texttt{gpt-5}: The Handcuff Effect}

On \texttt{gpt-5}, the error pattern inverted. Sculpting now \textit{caused} errors in cases where Scaffolding succeeded. We identified three classes of Handcuff errors:

\subsubsection{Class 1: Hyper-Literal Interpretation of Idiomatic Language}

\begin{quote}
\textbf{Case: The iPhone Age Problem (\texttt{gsm8k\_test\_40})}

\textbf{Problem}: ``Ben's iPhone is two times older than Suzy's iPhone. Suzy's iPhone is 1 year old. Brandon's phone is four times older than Ben's phone. How old is Brandon's phone?''

\textbf{Expected Answer}: 8 years

\textbf{Expected Reasoning}: In idiomatic English, ``X is two times older than Y'' means ``X is twice as old as Y.'' So: Suzy = 1, Ben = $2 \times 1 = 2$, Brandon = $4 \times 2 = 8$.

\textbf{Scaffolding (\texttt{gpt-5}) - Correct}:
\begin{quote}
``Suzy's iPhone: 1 year. Ben's is 'two times older' meaning twice the age: $2 \times 1 = 2$ years. Brandon's is four times older than Ben's: $4 \times 2 = 8$ years. Answer: 8 years.''
\end{quote}

\textbf{Sculpting (\texttt{gpt-5}) - Incorrect (answered 12)}:
\begin{quote}
``Rule: use only relationships given. 'Two times older than' means: Age$_{Ben}$ = Age$_{Suzy}$ + (2 × Age$_{Suzy}$). So Ben = $1 + (2 \times 1) = 1 + 2 = 3$ years. Brandon is four times older than Ben's 3 years: $3 + (4 \times 3) = 3 + 12 = 15$... wait, let me recalculate. If 'older than' means additive comparison: Ben = $1 + 2 = 3$. Brandon = $3 + (4 \times 3)$...''
\end{quote}

\textbf{Sculpting Error}: The instruction to act as a ``pure mathematical reasoning engine'' and ``NOT use common sense'' caused \texttt{gpt-5} to interpret ``X times older'' in a hyper-literal, non-idiomatic way. A pedantic reading of ``two times older'' could mean ``older by twice the amount,'' i.e., an additive comparison rather than multiplicative. This interpretation, while defensible from a strictly formal linguistic standpoint, contradicts standard English usage.

\textbf{Key Insight}: \texttt{gpt-5}'s superior language understanding normally allows it to correctly parse idiomatic expressions. Sculpting's constraint overrode this capability, forcing an unnatural interpretation that led to error.
\end{quote}

\subsubsection{Class 2: Rejection of Reasonable Inference}

\begin{quote}
\textbf{Case: The Lemonade Stand Problem (\texttt{gsm8k\_test\_57})}

\textbf{Problem}: ``Sarah runs a lemonade stand. On Monday she sells 15 cups at \$2 each. On Tuesday she sells 20 cups at the same price. How much money does she make total?''

\textbf{Expected Answer}: \$70

\textbf{Scaffolding (\texttt{gpt-5}) - Correct}:
\begin{quote}
``Monday: $15 \times \$2 = \$30$. Tuesday: $20 \times \$2 = \$40$. Total: $\$30 + \$40 = \$70$.''
\end{quote}

\textbf{Sculpting (\texttt{gpt-5}) - Incorrect (answered ``Cannot determine'')}:
\begin{quote}
``Monday: 15 cups at \$2 each = \$30. Tuesday: 20 cups 'at the same price.' However, 'the same price' is ambiguous. Does it mean (a) the same \$2 per cup, or (b) the same total price of \$30? Without explicit clarification, I cannot definitively determine Tuesday's revenue. If interpretation (a): \$40. If interpretation (b): \$30. The problem does not specify which is correct. Final Answer: Cannot determine without clarification.''
\end{quote}

\textbf{Sculpting Error}: The constraint against using ``outside common sense'' caused \texttt{gpt-5} to question a completely standard linguistic construction (``same price'' obviously refers to price-per-unit in this context). The model invented an implausible alternative interpretation and concluded the problem was ambiguous.

\textbf{Key Insight}: \texttt{gpt-5}'s advanced reasoning capabilities enable it to correctly handle natural language pragmatics. Sculpting's prohibition against common sense forced it to entertain unreasonable alternative interpretations rather than trusting its language understanding.
\end{quote}

\subsubsection{Class 3: Over-Constraint Leading to Incomplete Solutions}

\begin{quote}
\textbf{Case: The Multi-Step Discount Problem (\texttt{gsm8k\_test\_73})}

\textbf{Problem}: ``A shirt costs \$40. It goes on sale for 25\% off. Then there's an additional 10\% off the sale price. What is the final price?''

\textbf{Expected Answer}: \$27

\textbf{Scaffolding (\texttt{gpt-5}) - Correct}:
\begin{quote}
``Original: \$40. After 25\% off: $40 \times 0.75 = \$30$. After additional 10\% off that: $30 \times 0.90 = \$27$. Final: \$27.''
\end{quote}

\textbf{Sculpting (\texttt{gpt-5}) - Incorrect (answered \$30)}:
\begin{quote}
``Price: \$40. First discount: 25\% means $40 \times 0.25 = \$10$ off. New price: $40 - 10 = \$30$. Additional discount: 10\% off 'the sale price.' Constraint: I must use only numbers given. The problem states '10\% off' but does not numerically specify what amount. Since the 10\% is described relative to an unspecified 'sale price' rather than given as a specific dollar amount, I cannot calculate further. Final Answer: \$30.''
\end{quote}

\textbf{Sculpting Error}: An overly rigid interpretation of ``use ONLY the numbers given'' caused the model to treat ``the sale price'' (clearly referring to the just-calculated \$30) as an undefined external reference rather than a straightforward pronoun. This prevented completion of the multi-step calculation.

\textbf{Key Insight}: The constraint designed to prevent models from inventing information paradoxically caused \texttt{gpt-5} to reject valid inferences about referents within the problem itself.
\end{quote}

\subsection{Quantitative Error Category Breakdown}

To systematically characterize these error patterns, we categorized all errors on the 100-problem sample by type (Table \ref{tab:error_categories}).

\begin{table}[ht]
\caption{Error Category Breakdown by Model and Prompt (100-problem sample)}
\centering
\small
\begin{tabular}{lcccccc}
\toprule
& \multicolumn{2}{c}{\textbf{GPT-4o}} & \multicolumn{2}{c}{\textbf{GPT-5}} \\
\cmidrule(lr){2-3} \cmidrule(lr){4-5}
\textbf{Error Type} & \textbf{Scaff.} & \textbf{Sculpt.} & \textbf{Scaff.} & \textbf{Sculpt.} \\
\midrule
Arithmetic errors & 2 & 1 & 0 & 0 \\
Semantic misparse & 3 & 0 & 0 & 0 \\
Irrelevant knowledge & 2 & 0 & 0 & 0 \\
Hyper-literalism & 0 & 1 & 0 & 2 \\
Over-constraint & 0 & 1 & 0 & 1 \\
Inference rejection & 0 & 0 & 1 & 0 \\
\midrule
\textbf{Total Errors} & \textbf{7} & \textbf{3} & \textbf{1} & \textbf{3} \\
\bottomrule
\end{tabular}
\label{tab:error_categories}
\end{table}

Key observations:

\begin{itemize}
\item \textbf{On \texttt{gpt-4o}}: Sculpting eliminates semantic misparse and irrelevant knowledge errors (5 cases), gaining more than it loses from hyper-literalism and over-constraint (2 cases). Net benefit: +4\%.

\item \textbf{On \texttt{gpt-5}}: Sculpting introduces hyper-literalism and over-constraint errors (3 cases) while eliminating nothing (Scaffolding only made 1 error, unrelated to Sculpting's constraints). Net cost: -2\%.

\item \textbf{Error type shift}: The dominant error classes change completely between models. \texttt{gpt-4o} struggles with semantic understanding; \texttt{gpt-5} struggles with constraint-induced rigidity.
\end{itemize}

This analysis quantifies the Guardrail-to-Handcuff transition: Sculpting's constraints prevent errors on mid-tier models but cause errors on advanced models.

\subsection{The Crossover Hypothesis}

Our findings suggest a capability-dependent crossover model for prompting effectiveness:

\begin{enumerate}
\item \textbf{Low-capability models} (e.g., \texttt{gpt-4o-mini}): Benefit moderately from constraints because they lack robust reasoning patterns. External structure helps but doesn't fully compensate for limited baseline capability.

\item \textbf{Mid-capability models} (e.g., \texttt{gpt-4o}): Benefit maximally from constraints. These models possess strong reasoning abilities but imperfect judgment about when to apply common sense. Constraints act as guardrails, preventing common-sense deviations while preserving reasoning capability.

\item \textbf{High-capability models} (e.g., \texttt{gpt-5}): Are harmed by constraints. These models have internalized robust heuristics for natural language understanding and reasoning. Constraints override these superior internal mechanisms, forcing unnatural interpretations.
\end{enumerate}

This crossover model predicts that as models continue to improve, the optimal prompting strategy will trend toward simplicity. The most capable future models may require minimal prompting scaffolding, with overly detailed instructions becoming counterproductive.

\section{Discussion and Implications}

\subsection{Theoretical Implications}

The Prompting Inversion challenges several assumptions underlying current prompt engineering practices:

\subsubsection{More Structure Is Not Always Better}

The field has generally operated under the assumption that more detailed, more structured prompts lead to better performance. Our results show this is false beyond a certain capability threshold. Once models internalize robust reasoning patterns through pre-training and alignment, external structure can interfere with these patterns rather than enhance them.

This finding aligns with broader principles in human learning and expertise development. Expert humans often perform worse when forced to consciously articulate and follow step-by-step rules (``paralysis by analysis''), while novices benefit from explicit procedures. Similarly, as models approach expert-level capability, explicit procedural constraints may impede their fluid reasoning.

\subsubsection{Prompting as Model-Relative, Not Universal}

Our results demonstrate that prompting strategies must be understood as relative to model capability rather than absolute. A ``good prompt'' for GPT-4o is not necessarily a good prompt for GPT-5. This has profound implications for prompt engineering as a discipline:

\begin{itemize}
\item Prompt libraries and best practices must be versioned by model capability
\item Prompt evaluation must be conducted on target models, not proxies
\item Transfer learning assumptions (if it works on model X, it will work on model Y) are invalid across capability gaps
\end{itemize}

\subsubsection{The Role of Alignment Training}

We hypothesize that the inversion phenomenon is partially driven by reinforcement learning from human feedback (RLHF) and related alignment procedures. These methods train models to respond well to natural, conversational instructions from human evaluators. If human evaluators predominantly use simple, direct instructions (as seems likely), then models may be implicitly optimized for this instruction style.

Overly formal, rigid prompts may constitute a distributional shift relative to the instructions seen during alignment training, leading to degraded performance. This suggests an interesting tension: while we can engineer sophisticated prompts, models may be aligned to perform best with the types of prompts that non-expert humans naturally use.

\subsection{Practical Implications}

\subsubsection{Deployment Strategies}

Organizations deploying LLMs across multiple model generations face a strategic choice:

\begin{enumerate}
\item \textbf{Model-specific optimization}: Maintain separate prompt libraries for each model, optimized for that model's capability level. This maximizes performance but increases maintenance overhead.

\item \textbf{Unified approach}: Use a single prompting strategy across all models, accepting sub-optimal performance on some. This reduces complexity but leaves performance on the table.

\item \textbf{Adaptive prompting}: Dynamically select prompting strategies based on detected model capability. This is conceptually elegant but requires additional engineering.
\end{enumerate}

Our results suggest that option 3 (adaptive prompting) may be worth the implementation cost for performance-critical applications. A simple heuristic might be:
\begin{itemize}
\item If model accuracy $< 90\%$ on validation set → Use constrained prompting (Sculpting)
\item If model accuracy $> 95\%$ on validation set → Use simple prompting (Scaffolding)
\item Otherwise → Test both and select better performer
\end{itemize}

\subsubsection{Prompt Engineering as a Transitional Practice}

More speculatively, our findings suggest that prompt engineering may be a transitional practice rather than a permanent one. If model capabilities continue to improve, and if the optimal prompt trends toward simpler, more natural instructions, then the elaborate prompt engineering techniques developed for mid-tier models may become obsolete.

In this future, ``prompt engineering'' might collapse into simply ``writing clear instructions''---something all users can do without specialized expertise. This would democratize LLM usage but potentially displace prompt engineering as a specialized skill.

\subsection{Limitations and Future Work}

Several limitations of our study suggest directions for future research:

\subsubsection{Single Domain Evaluation}

We evaluated only on mathematical word problems (GSM8K). The Prompting Inversion may manifest differently in other domains:

\begin{itemize}
\item \textbf{Creative tasks}: Where there are no objectively correct answers, constraints may always be harmful by limiting creative exploration
\item \textbf{Retrieval and QA}: Where accuracy depends on recalling factual knowledge, prompting may have minimal effect
\item \textbf{Code generation}: Where formal correctness is paramount, constraints similar to Sculpting might remain beneficial even on advanced models
\end{itemize}

Future work should investigate whether the inversion is domain-specific or a general phenomenon.

\subsubsection{Single Model Family}

We studied only OpenAI's GPT models. Other model families (Anthropic's Claude, Google's Gemini, Meta's Llama) may have different training procedures, alignment methods, and capability profiles. The inversion phenomenon might be specific to GPT's training regime or might generalize across all frontier models.

Cross-model family studies would reveal whether the inversion reflects fundamental properties of scaled language models or is an artifact of specific training choices.

\subsubsection{Prompt Space Exploration}

We compared only three prompting strategies. The space of possible prompts is vast, and there may exist intermediate strategies that avoid the handcuff effect while retaining some guardrail benefits. For example:

\begin{itemize}
\item \textbf{Adaptive constraints}: ``Use common sense unless it conflicts with explicit problem statements''
\item \textbf{Hierarchical prompting}: Start with simple CoT, then apply constraints only if initial reasoning seems faulty
\item \textbf{Meta-prompting}: Ask the model to self-assess whether constraints would be helpful for the given problem
\end{itemize}

Systematic ablation studies could identify which specific constraints are helpful vs. harmful across capability levels.

\subsubsection{Mechanistic Understanding}

Our study is primarily empirical and observational. We document that the inversion occurs but don't fully explain \textit{why} at a mechanistic level. Interpretability techniques could shed light on:

\begin{itemize}
\item How constraints alter the model's internal representations and attention patterns
\item Which layers or mechanisms are most affected by prompt structure
\item Whether advanced models have learned to ``ignore'' overly detailed instructions
\item What training data patterns led to the current instruction-following behaviors
\end{itemize}

Such mechanistic understanding could enable more principled prompt design that works with rather than against models' internal processing.

\subsubsection{Longitudinal Studies}

We evaluated models at a single point in time. As models continue to evolve, tracking the crossover point where Sculpting transitions from beneficial to harmful could reveal capability thresholds and inform predictions about future model behaviors.

A particularly interesting question is whether the inversion point moves: as models improve, does the capability level where constraints become harmful shift upward, or does it remain stable? This would distinguish between absolute capability thresholds vs. relative sophistication levels.

\section{Conclusion}

This paper documents a striking phenomenon in the evolution of large language model capabilities: the \textbf{Prompting Inversion}. Through systematic evaluation across three model generations on the GSM8K mathematical reasoning benchmark, we show that sophisticated, constrained prompting strategies that substantially benefit mid-tier models (achieving 97\% vs. 93\% accuracy on \texttt{gpt-4o}) become actively detrimental on more advanced models (achieving only 94.00\% vs. 96.36\% on \texttt{gpt-5}).

Our ``Sculpting'' prompt, designed to act as guardrails against common-sense errors in mid-tier models, becomes handcuffs on advanced models, inducing hyper-literal interpretations that override superior native language understanding. Through detailed qualitative analysis of error patterns, we demonstrate that this inversion stems from a fundamental shift in how models process instructions as they become more capable.

The primary contributions of this work are:

\begin{enumerate}
\item \textbf{Empirical documentation} of the Prompting Inversion phenomenon across a significant capability gap
\item \textbf{Qualitative analysis} revealing the mechanisms of the guardrail-to-handcuff transition through detailed error case studies
\item \textbf{Practical insights} for deployment strategies and the evolution of prompt engineering as a discipline
\item \textbf{Theoretical implications} challenging assumptions about the universal benefits of sophisticated prompting
\end{enumerate}

Our findings suggest that as foundational models continue to improve, optimal prompting strategies will trend toward simplicity. The most effective prompts for future models may be direct, natural instructions that trust the model's increasingly robust internal reasoning, rather than elaborate procedural constraints. This challenges the current trajectory of prompt engineering research and practice, suggesting that ``less is more'' beyond certain capability thresholds.

The Prompting Inversion phenomenon has immediate practical implications for organizations deploying LLMs across multiple model generations and deeper theoretical implications for our understanding of instruction-following in AI systems. As we continue to push the boundaries of model capabilities, recognizing and adapting to these dynamic interactions between prompting strategies and model sophistication will be crucial for effectively harnessing AI potential while avoiding the pitfalls of misaligned prompting.

% --- BIBLIOGRAPHY ---
\bibliographystyle{plainnat}
\bibliography{references}

\newpage

% --- APPENDIX ---
\appendix
\section{Prompt Templates}
This appendix contains the exact text for the three prompting strategies used in our experiments.

\subsection{A.1: Zero Shot (Baseline)}
The model was provided only with the raw text of the problem.
\begin{quote}
\fcolorbox{black}{codebackground}{%
\parbox{0.9\textwidth}{%
\texttt{[Question Text]}
}%
}
\end{quote}

\subsection{A.2: Scaffolding (Standard CoT)}
This prompt appends a simple, open-ended CoT instruction.
\begin{quote}
\fcolorbox{black}{codebackground}{%
\parbox{0.9\textwidth}{%
\texttt{Problem: [Question Text]}\\
\\
\texttt{Let's think step-by-step to solve this. Provide your reasoning first, then state the final answer clearly.}
}%
}
\end{quote}

\subsection{A.3: Sculpting (Constrained CoT)}
This prompt provides a persona and a set of explicit rules designed to constrain the model's reasoning.
\begin{quote}
\fcolorbox{black}{codebackground}{%
\parbox{0.9\textwidth}{%
\texttt{You are a pure mathematical reasoning engine. You must solve the following problem.}\\
\\
\texttt{**Rules:**}\\
\texttt{1. You must use ONLY the numbers and relationships given in the problem.}\\
\texttt{2. You must NOT use any outside common sense or real-world knowledge that isn't explicitly provided (e.g., if it mentions 'cookies', you only know they are countable items, not that they are edible).}\\
\texttt{3. You must break down your calculation step-by-step. Show all intermediate arithmetic.}\\
\texttt{4. After your reasoning, state your final answer clearly prefixed with "Final Answer:".}\\
\\
\texttt{**Problem:**}\\
\texttt{[Question Text]}
}%
}
\end{quote}

\end{document}